%% file: main.tex
\documentclass[letterpaper,10pt,conference]{ieeeconf}
\IEEEoverridecommandlockouts
\usepackage{cite}
\usepackage{amsmath,amssymb}
\usepackage{graphicx,booktabs,array}
\newcommand{\statcell}[2]{$#1 \pm #2$}
\usepackage{xcolor,tikz}
\usetikzlibrary{arrows.meta,positioning}
\usepackage{dblfloatfix,flafter,placeins}
\usepackage[hidelinks]{hyperref}
\hypersetup{pdfauthor={},pdftitle={},pdfsubject={},pdfkeywords={}}
\title{\LARGE\bf Learning Grasp Targeting from Point Clouds\\for Log Pile Clearing on a Hydraulic Crane}
\author{George Sideris$^{1}$, Lucas Bessai$^{1}$, Heshan Fernando$^{2}$,\\
Elie Ayoub$^{2}$, Nicolas Lemieux$^{2}$ and Inna Sharf$^{1}$%
\thanks{$^{1}$McGill University, Montreal, QC, Canada.}%
\thanks{$^{2}$FPInnovations, Pointe-Claire, QC, Canada.}%
\thanks{Supplementary video: \url{https://youtu.be/woyDNW_oN7A}.}}
\IEEEaftertitletext{\vspace{-\baselineskip}}
\makeatletter
\long\def\@makecaption#1#2{%
\setbox\@tempboxa\hbox{\footnotesize #1.~~ #2}%
\ifx\@captype\@IEEEtablestring\else\@IEEEfigurecaptionsepspace\fi
\ifdim \wd\@tempboxa >\hsize%
\setbox\@tempboxa\hbox{\footnotesize #1.~~ }%
\parbox[t]{\hsize}{\footnotesize \noindent\unhbox\@tempboxa#2}%
\else%
\ifcenterfigcaptions \hbox to\hsize{\footnotesize\hfil\box\@tempboxa\hfil}%
\else \hbox to\hsize{\footnotesize\box\@tempboxa\hfil}%
\fi\fi
\ifx\@captype\@IEEEtablestring\@IEEEtablecaptionsepspace\fi}
\makeatother
\begin{document}
\maketitle
\thispagestyle{empty}
\pagestyle{empty}

\begin{abstract}
In mill yards, log loaders clear dense piles by a sequence of bundle grasps: hundreds of logs rest in contact, and each removal changes the pile available to the next grasp. A learned policy chooses where to place and orient the grapple from unsegmented point clouds and runs on a trailer-mounted hydraulic forestry crane. The policy classifies at which observed point to grasp and predicts depth and grapple orientation there. The same network outputs support behavior cloning (BC), reinforcement learning (RL), and deployment. BC learns from successful top-of-pile demonstrations; RL explores for improvements by fine-tuning the cloned policy (BC$\to$RL) or by training from scratch. In simulation, BC clears 98 of 100 piles of 200 logs, while BC$\to$RL improves load stability. Twelve field trials compare a geometric heuristic, RL from scratch, BC, and BC$\to$RL through complete grasp-transport-deposit cycles. BC and BC$\to$RL deposit 93.8\% and 88.9\% of pooled inventory, against 80.4\% for the heuristic. BC$\to$RL deposits logs on 83.6\% of its cycles, against 79.6\% for the heuristic and 65.7\% for BC, while its simulated stability gain does not carry over to the crane testbed. Trained entirely in simulation and run unchanged on the crane, the learned policies clear more than the hand-filtered heuristic while observing unfiltered clouds that still contain the storage rack's rails and poles.
\end{abstract}

\section{INTRODUCTION}
Log handling is repetitive work in forestry and mill yards, where automation could ease operator shortages and reduce exposure to unstable loads~\cite{lindroos2017forestry,visser2021automation}. A grapple-equipped crane repeatedly grasps and lifts bundles from a dense pile, in which hundreds of logs rest against one another, then carries them to the mill's infeed deck. Each such grasp, carry, and deposit forms a \emph{cycle}, and every cycle changes the scene for the next decision: closing the grapple moves the captured bundle and the surrounding material, and as the pile depletes, broad mounds give way to isolated logs beside rails and poles.

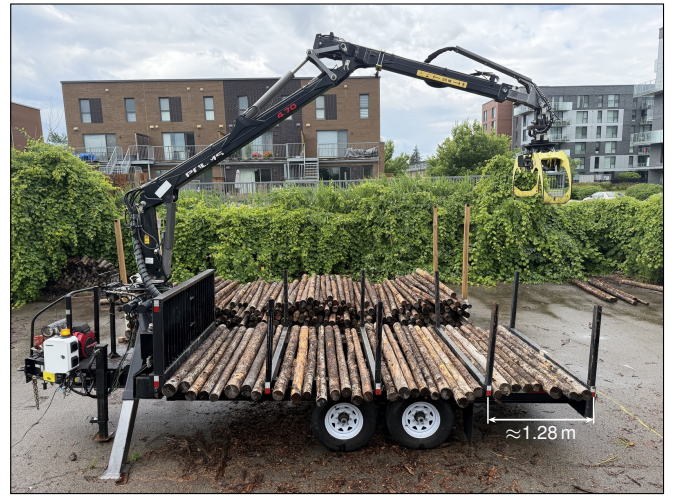
\begin{figure}[!b]
\centering
\begin{tikzpicture}
\node[anchor=south west,inner sep=0,outer sep=0] (photo) at (0,0)
  {\includegraphics[width=\columnwidth]{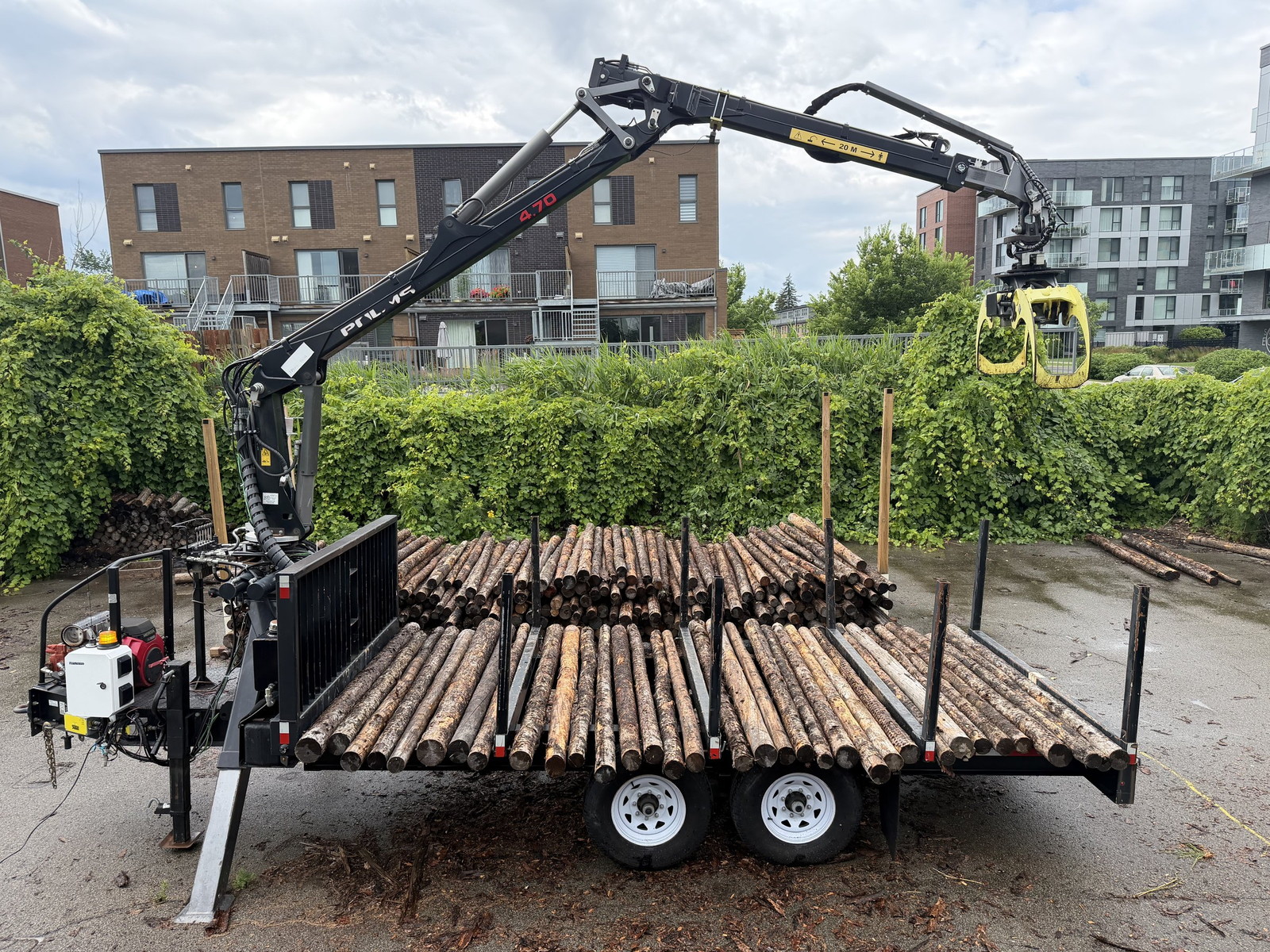}};
\begin{scope}[x={(photo.south east)},y={(photo.north west)}]
\draw[white,line width=0.55pt]
  (0.731,0.197)--(0.731,0.143)
  (0.893,0.197)--(0.893,0.143);
\draw[{Stealth[length=3pt]}-{Stealth[length=3pt]},white,line width=0.65pt]
  (0.731,0.151)--(0.893,0.151)
  node[midway,below=1pt,inner sep=1pt,text=white,font=\sffamily\scriptsize]
  {$\approx$1.28\,m};
\end{scope}
\draw[black,line width=0.4pt] (photo.south west) rectangle (photo.north east);
\end{tikzpicture}
\caption{Trailer-mounted hydraulic forestry crane, suspended grapple, source rack, and destination trailer. The policy selects the grasp; a common controller transports and deposits the logs.}
\label{fig:system}
\end{figure}

The decision studied here is where to place and orient the grapple, given a partial, noisy point cloud of the pile. The learned policy runs on a trailer-mounted hydraulic forestry crane with a passively suspended grapple (Fig.~\ref{fig:system}). A common controller executes the resulting grasp, transport, and deposition; learning concerns only the grasp decision.

A top-of-pile rule is the natural start, but a camera cloud requires distinguishing logs from rack structure and stray returns; geometric filters depend on scene geometry and calibration, and a segmentation network needs its own labels. Learning the cloud-to-grasp mapping trains material selection and grasp prediction together, without segmentation.

Behavior cloning (BC) learns from successful grasps of a simulated expert that knows the log poses and targets the highest log. Reinforcement learning (RL) explores for improvements: fine-tuning seeks better bundle handling, while RL from scratch explores for more effective grasp sequences beyond the expert's top-of-pile strategy.

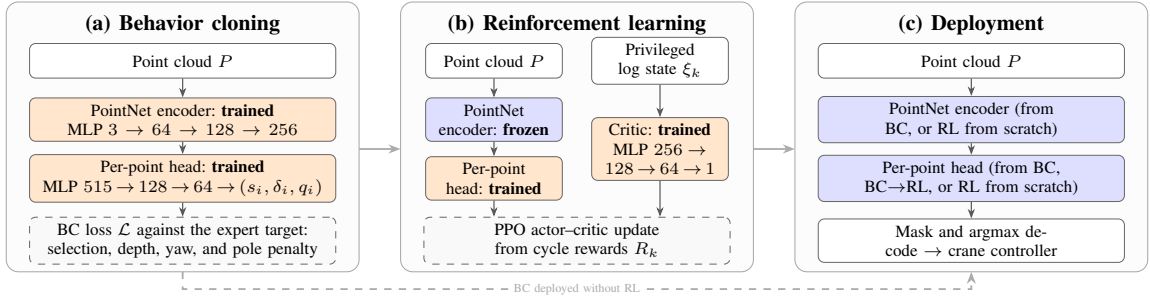
\begin{figure*}[t]
\centering
\resizebox{0.85\textwidth}{!}{%
\begin{tikzpicture}[
  box/.style={rectangle,rounded corners=2.5pt,draw=black!60,align=center,
    font=\scriptsize,text width=4.25cm,minimum height=0.50cm,inner sep=2.5pt},
  train/.style={box,fill=orange!20},
  frozen/.style={box,fill=blue!13},
  io/.style={box,fill=white},
  opt/.style={box,fill=gray!5,dashed},
  flow/.style={-{Stealth[length=4pt]},line width=0.5pt,black!65},
  transfer/.style={-{Stealth[length=5pt]},line width=0.75pt,gray!65}]

\draw[rounded corners=5pt,draw=black!35,fill=gray!2]
  (-2.55,0.55) rectangle (2.55,-3.35);
\draw[rounded corners=5pt,draw=black!35,fill=gray!2]
  (3.15,0.55) rectangle (8.25,-3.35);
\draw[rounded corners=5pt,draw=black!35,fill=gray!2]
  (8.85,0.55) rectangle (13.95,-3.35);
\node[font=\small\bfseries] at (0,0.22) {(a) Behavior cloning};
\node[font=\small\bfseries] at (5.7,0.22) {(b) Reinforcement learning};
\node[font=\small\bfseries] at (11.4,0.22) {(c) Deployment};

\node[io] (bcin) at (0,-0.30) {Point cloud $P$};
\node[train] (bcenc) at (0,-1.15) {PointNet encoder: \textbf{trained}\\MLP $3\!\to\!64\!\to\!128\!\to\!256$};
\node[train] (bchead) at (0,-2.00) {Per-point head: \textbf{trained}\\MLP $515\!\to\!128\!\to\!64\!\to\!(s_i,\delta_i,q_i)$};
\node[opt] (bcloss) at (0,-2.90) {BC loss $\mathcal L$ against the expert target:\ selection, depth, yaw, and pole penalty};
\draw[flow] (bcin)--(bcenc); \draw[flow] (bcenc)--(bchead); \draw[flow] (bchead)--(bcloss);

\node[io,text width=1.80cm] (rlin) at (4.50,-0.30) {Point cloud $P$};
\node[io,text width=1.80cm] (rlstate) at (6.90,-0.30) {Privileged log state $\xi_k$};
\node[frozen,text width=1.80cm] (rlenc) at (4.50,-1.15) {PointNet encoder: \textbf{frozen}};
\node[train,text width=1.80cm] (rlhead) at (4.50,-2.00) {Per-point head: \textbf{trained}};
\node[train,text width=1.80cm] (rlcritic) at (6.90,-1.575) {Critic: \textbf{trained}\\MLP $256\!\to\!128\!\to\!64\!\to\!1$};
\node[opt] (ppo) at (5.7,-2.90) {PPO actor--critic update from cycle rewards $R_k$};
\draw[flow] (rlin)--(rlenc); \draw[flow] (rlenc)--(rlhead);
\draw[flow] (rlhead.south)--(rlhead.south |- ppo.north);
\draw[flow] (rlstate)--(rlcritic);
\draw[flow] (rlcritic.south)--(rlcritic.south |- ppo.north);

\node[io] (depin) at (11.4,-0.30) {Point cloud $P$};
\node[frozen] (depenc) at (11.4,-1.15) {PointNet encoder (from BC, or RL from scratch)};
\node[frozen] (dephead) at (11.4,-2.00) {Per-point head (from BC, BC$\to$RL, or RL from scratch)};
\node[io] (decode) at (11.4,-2.90) {Mask and argmax decode $\to$ crane controller};
\draw[flow] (depin)--(depenc); \draw[flow] (depenc)--(dephead); \draw[flow] (dephead)--(decode);

\draw[transfer] (2.55,-1.575)--(3.15,-1.575);
\draw[transfer] (8.25,-1.575)--(8.85,-1.575);
\draw[transfer,dashed] (0,-3.35)--(0,-3.60)--node[midway,fill=white,inner sep=1pt,font=\tiny]{BC deployed without RL} (11.4,-3.60)--(11.4,-3.35);
\end{tikzpicture}}
\caption{Shared scoring policy through BC, RL, and deployment. Orange blocks are trained; blue blocks carry weights unchanged. Gray arrows carry weights between phases. The RL panel shows fine-tuning from BC; the dashed arrow is BC deployed directly, skipping (b). The privileged state $\xi_k$ contains the positions and yaws of up to 64 logs and is available only to the critic during training. RL from scratch also trains the encoder, enters (b) without BC weights, and deploys both of its blocks. MLP: multilayer perceptron; PPO: proximal policy optimization.}
\label{fig:architecture}
\end{figure*}

The contributions of this paper are:
\begin{itemize}
\item A learned grasp policy for sequential clearing of packed log piles from unsegmented point clouds. It classifies at which observed point to grasp and predicts the depth and the yaw, the grapple's rotation about the vertical axis, at that point; one network serves BC, RL fine-tuning, and deployment (Fig.~\ref{fig:architecture}). The point selection matters: a regression head trained on the same demonstrations fails to clear the piles (Sec.~\ref{sec:heads}).
\item Twelve field trials on the crane testbed in which BC and BC$\to$RL, trained entirely in simulation and run with unchanged weights on unfiltered clouds, deposit more inventory than the deployed geometric heuristic on two of three pile shapes and overall.
\item A comparison of the geometric heuristic, RL from scratch, BC, and BC$\to$RL in simulation and on the crane testbed showing that the learned policies' clearing advantage carries over while fine-tuning's handling gain does not, with field failures classified by cause and trial phase.
\end{itemize}

\section{RELATED WORK}
Forestry automation includes crane motion planning~\cite{ortizmorales2014increasing,jebellat2025designing}, learned control~\cite{andersson2021reinforcement}, vision-guided single-log grasping~\cite{semberg2024}, and loading by an unmanned forwarder~\cite{lahera2024feasibility}. Learned grasp planning has also been evaluated for small log clusters on a real loader~\cite{ayoub2023grasp}, for groups of two to five logs~\cite{wallin2024multilog}, and for clusters of up to seven~\cite{falldin2025synthesizing}. These works automate motion for a given target or plan one grasp on a static scene; none addresses sequential clearing of a pile whose state evolves under its own actions or reports pile-scale hardware trials.

Learned grasping addresses isolated objects and clutter~\cite{morrison2018closing,mahler2017dexnet}. Sequential bin picking~\cite{mahler2017sequential} and push-grasp policies~\cite{zeng2018learning} explicitly account for changes induced by earlier actions. That literature assumes loose, separable clutter and pinch grasps that singulate one object; mill-yard logs are packed in layers, nothing can be singulated, and the grapple takes bunches in one power grasp that reshapes the pile.

PointNet~\cite{qi2017pointnet} provides per-point and global cloud features; point-cloud policies support imitation~\cite{ze2024dp3} and simulation-to-real RL~\cite{qin2022dexpoint}, and HACMan~\cite{zhou2023hacman} selects point-cloud contact locations with continuous motion parameters for non-prehensile manipulation. Demonstration-initialized RL~\cite{rajeswaran2018learning}, proximal policy optimization (PPO)~\cite{schulman2017ppo}, and critics with privileged observations~\cite{pinto2018asymmetric} are the learning tools.

\section{METHOD}
\label{sec:method}
\subsection{Grasp representation}
\label{sec:representation}
At each cycle, indexed by $k$, the policy receives a point cloud $P=\{p_i\}$, where $p_i$ is the three-dimensional position of observed point $i$. It chooses a grasp position $t=(x,y,z)$ and yaw $\psi$ in the crane base frame. A fixed crane controller executes the target; the next cloud shows the changed pile.

A pile can present several separated graspable regions. Regressing a single coordinate with a squared-error loss can average those alternatives into a gap. Classifying over observed points preserves separate candidates. Section~\ref{sec:heads} tests this choice against a regression head on the same data and backbone.

A PointNet encoder applies a shared multilayer perceptron (MLP), $3\to64\to128\to256$ with batch normalization and an exponential linear unit after each layer, to every point, giving the per-point feature $h_i$. Max pooling over points and a $256\to256$ projection with the same normalization and activation produce the global feature $g$. A second shared MLP reads, for every point, the concatenation $[h_i;g;p_i]$ of its 256-wide feature, the 256-wide global feature, and its position, 515 values in all, and three linear outputs on its last layer predict, at that point, selection score $s_i$, vertical offset $\delta_i$, and doubled-angle yaw encoding $q_i=(q_i^c,q_i^s)$. Figure~\ref{fig:architecture} gives all actor widths and outputs. The head input has dropout probability 0.2 during BC training.

\begin{figure*}[t]
\centering
\includegraphics[width=0.325\textwidth]{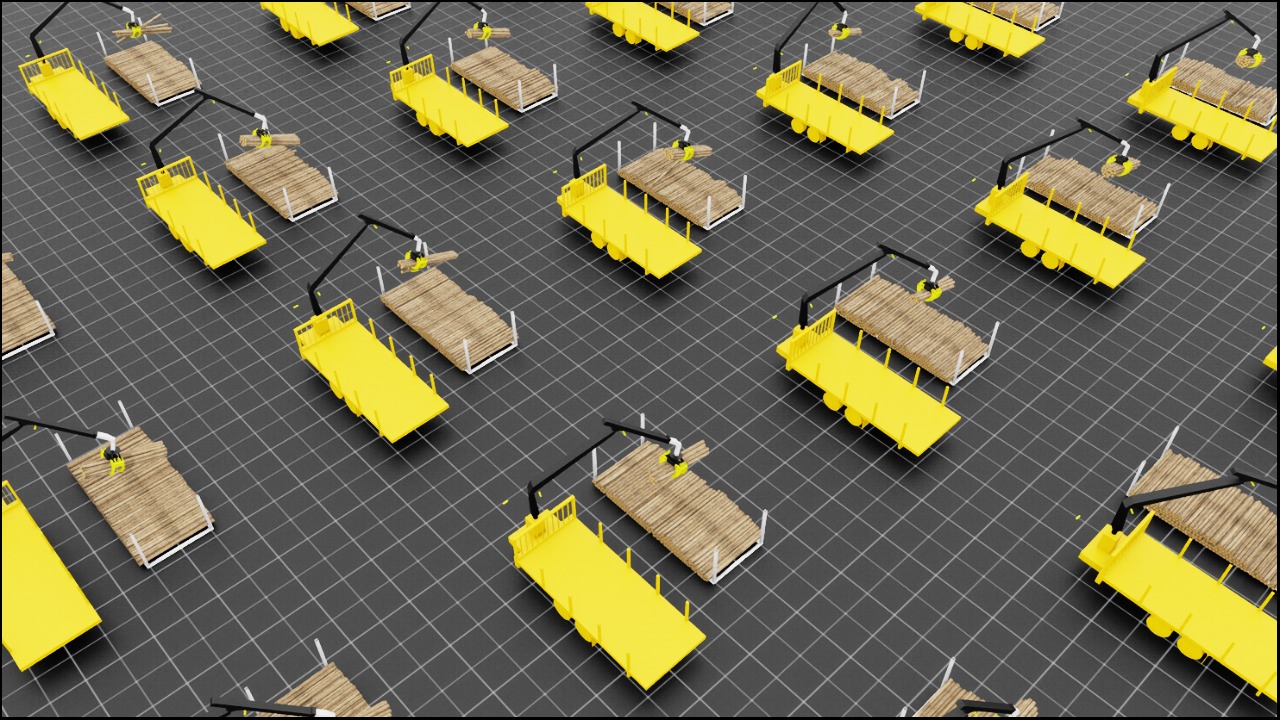}\hfill
\includegraphics[width=0.325\textwidth]{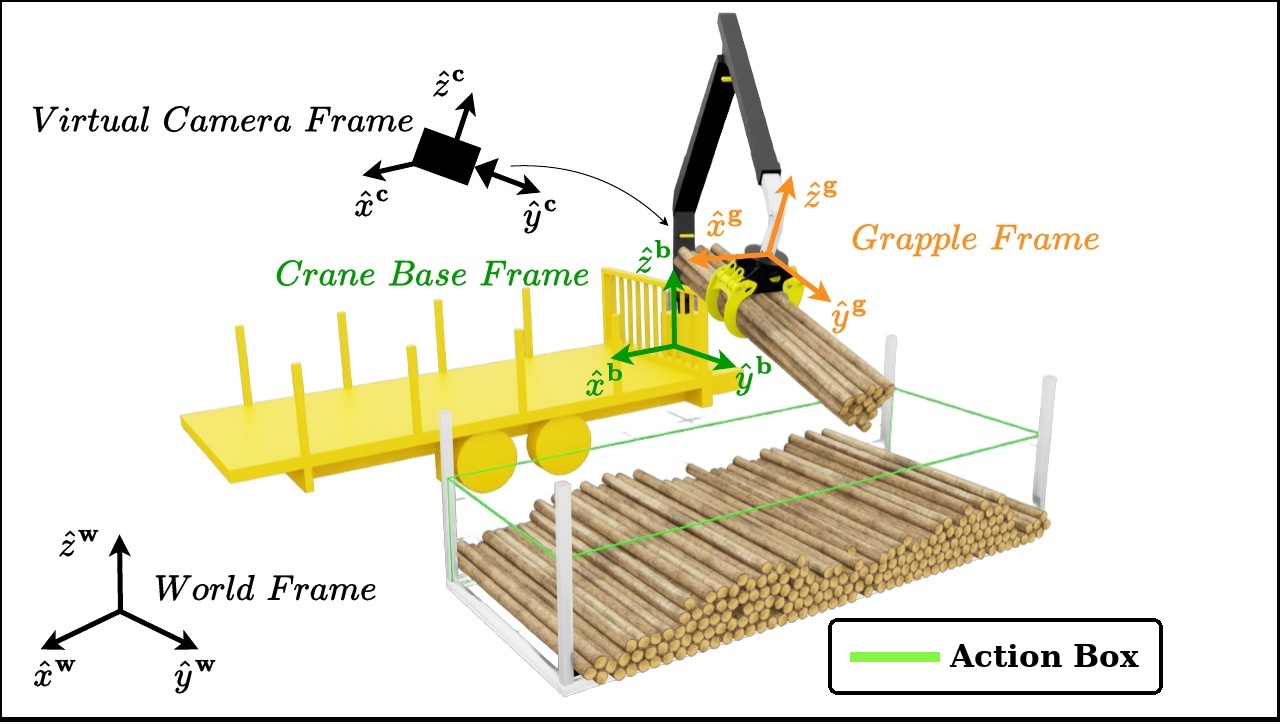}\hfill
\includegraphics[width=0.325\textwidth]{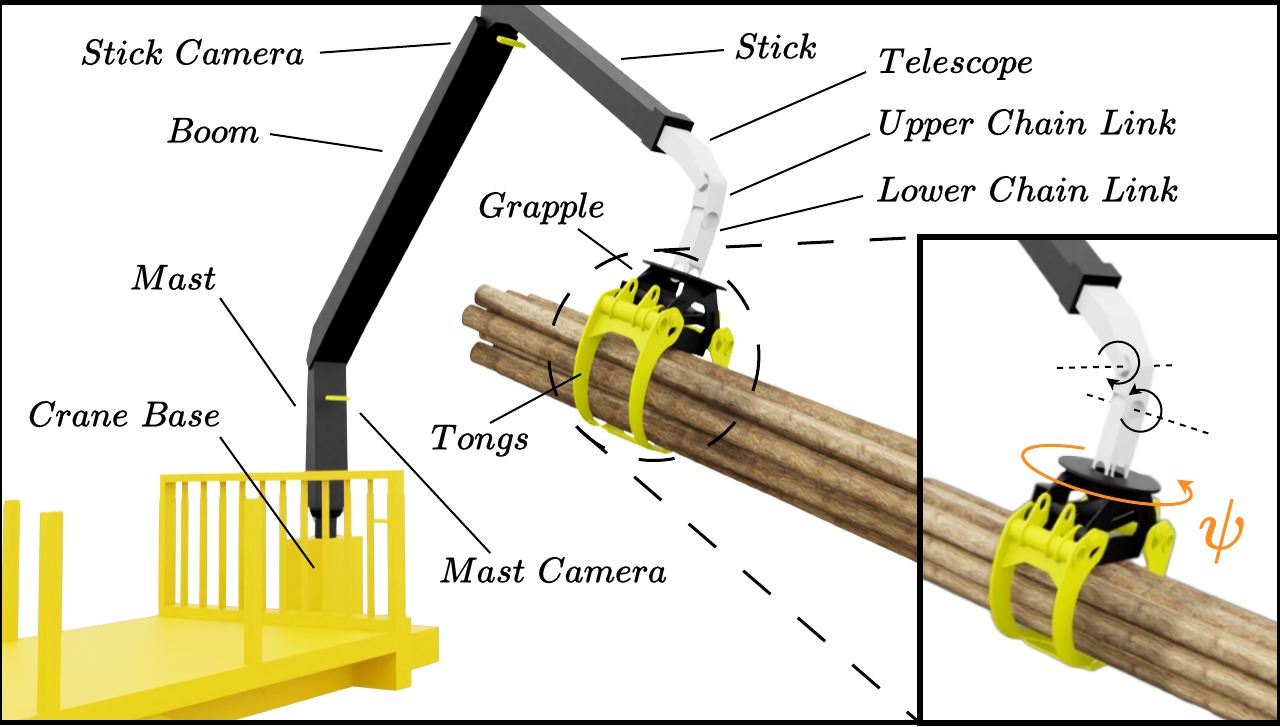}
\caption{Simulation: parallel environments (left), rack and frames (middle), and actuated links with passive suspension (right). $\psi$ is grapple yaw.}
\label{fig:sim_env}
\end{figure*}

\begin{figure}[t]
\centering
\includegraphics[width=\columnwidth]{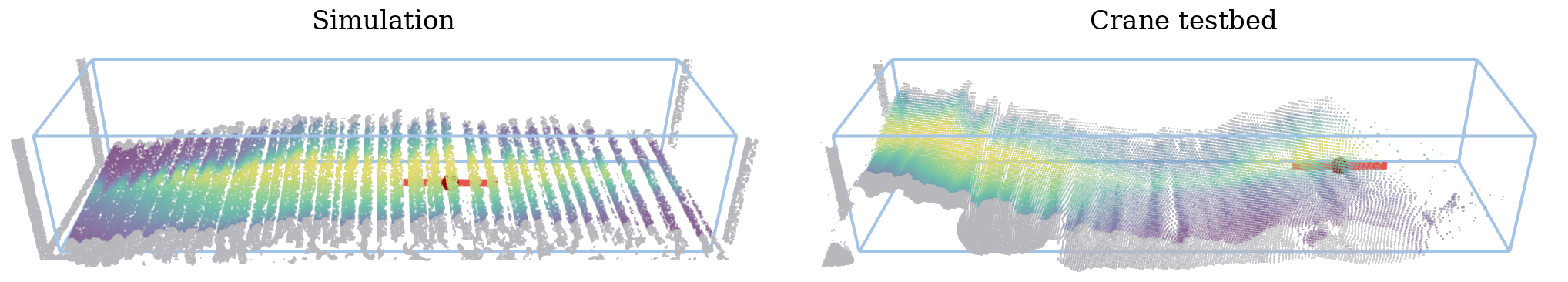}
\caption{Scores on simulated (left) and crane (right) clouds. Brightness gives score rank, gray marks points outside the action box, and red gives the decoded target and yaw.}
\label{fig:scores}
\end{figure}

Targets are restricted to a predefined grasping volume within the source rack, called the \emph{action box} (Fig.~\ref{fig:sim_env}, middle). Padded input slots and points outside that volume are masked. At deployment, the highest-scoring unmasked point, with index $\hat\imath$, determines the target position $\hat t$ and yaw $\hat\psi$:
\begin{equation}
\begin{aligned}
 \hat\imath&=\arg\max_i s_i,\\
 \hat t&=(p_{\hat\imath,x},p_{\hat\imath,y},p_{\hat\imath,z}+\delta_{\hat\imath}),\\
 \hat\psi&=\tfrac12\operatorname{atan2}(q_{\hat\imath}^s,q_{\hat\imath}^c).
\end{aligned}
\label{eq:decode}
\end{equation}
The doubled-angle yaw respects the equivalence of grasps separated by half a turn~\cite{morrison2018closing}. The vertical offset allows the target to lie below the selected return. Selecting among observed points restricts horizontal targets to measured surfaces, which can include rack structure and noise. The policy must therefore learn which material is graspable (Fig.~\ref{fig:scores}).

\subsection{Behavior cloning}
\label{sec:bc}
A privileged expert selects the highest simulated log and aligns the grapple with its axis. Its successful cycles supply 11,366 cloud--target demonstrations from 500 simulated pile-clearing sequences, or \emph{episodes}. Two commissioning probes on the crane found that surface-level commands raked the pile, while targets 0.25--0.30~m below the surface closed on material. The recorded targets $t$ are log centers; their height $t_z$ is relabeled to the training label $t_z'$ with depth $d=0.25$~m:
\begin{equation}
 t_z'=\max(t_z+r_{\log}-d,\;z_{\mathrm{label}}),
 \label{eq:depth}
\end{equation}
where $r_{\log}=0.056$~m estimates the surface above the log center and $z_{\mathrm{label}}=-1.20$~m is the lowest label allowed. These relabeled depths train BC and initialize its RL fine-tuning, BC$\to$RL; RL from scratch never sees them.

Let $i^*$ index the point horizontally nearest the expert target. The target distribution assigns equal probability $y_i$ to points within 0.12~m of the target, always including $i^*$, and zero elsewhere. The BC loss combines point selection, depth, and yaw errors with a rack-pole penalty:
\begin{equation}
\begin{aligned}
 \mathcal L={}&-\sum_i y_i\log\operatorname{softmax}(s)_i
       +\mathcal L_\delta+0.5\mathcal L_q+\mathcal L_{\mathrm{neg}},\\
 \mathcal L_\delta={}&\big(\delta_{i^*}-(t_z'-p_{i^*,z})\big)^2,\\
 \mathcal L_q={}&\|q_{i^*}-(\cos2\psi,\sin2\psi)\|_2^2.
\end{aligned}
\label{eq:loss}
\end{equation}
Pole columns are shortened randomly, and $\mathcal L_{\mathrm{neg}}$ averages $\log(1+e^{s_i})$ over their remaining points to discourage selection of visible rack poles. Joint horizontal translations move cloud and target together. Camera-ray noise uses standard deviation $\sigma(r)=0.0014r^2$~m at range $r$ in meters~\cite{ortiz2018zed}.

BC uses AdamW with initial learning rate and weight decay $10^{-4}$, batch size 64, and gradient-norm clipping at 1.0. The learning rate halves when the loss on a held-out 15\% validation split of the demonstrations plateaus; early stopping retains the saved network weights, or \emph{checkpoint}, with the lowest validation loss.

\subsection{Reinforcement learning}
\label{sec:rl}
The expert selects the highest log without optimizing the resulting bundle. RL explores using a reward that values both the number of logs and their handling. During training, policy $\pi$ samples point index $i$, depth $\delta$, and yaw encoding $q$ from cloud $P$:
\begin{equation}
\begin{aligned}
 \pi(i,\delta,q\mid P)={}&\operatorname{Cat}(i\mid\operatorname{softmax}(s/\tau))\\
 &\cdot\mathcal N(\delta\mid\delta_i,\sigma_\delta^2)
 \mathcal N(q\mid q_i,\sigma_q^2 I_2).
\end{aligned}
\label{eq:policy}
\end{equation}
Temperature $\tau=1.0$ sets the score distribution; learned $\sigma_\delta$ and $\sigma_q$ control depth and yaw exploration, and $I_2$ is the identity matrix. Sampling a point concentrates horizontal exploration on observed surfaces; Section~\ref{sec:heads} compares this with a Gaussian policy over grasp coordinates, which can explore anywhere in the box.

The cycle reward $R$ combines the number of logs held in the grapple after lifting, $n$, their alignment score relative to the tong axis, $\alpha$, and grapple stability, $\varsigma$:
\begin{equation}
 R=\begin{cases}n\alpha\varsigma,&n>0,\\-1,&n=0.\end{cases}
 \label{eq:reward}
\end{equation}
Both handling scores are defined in Sec.~\ref{sec:outcomes}. Aligned, level loads earn more; an empty cycle is charged.

Each grasp affects later rewards by changing the pile, and PPO maximizes the return discounted with $\gamma=0.99$ over an episode of at most $H=30$ cycles. An asymmetric critic estimates remaining return from privileged state $\xi_k\in\mathbb R^{256}$, which contains the positions and yaws of up to the 64 highest logs, sorted by height. The actor sees only the point cloud, while the critic receives the privileged state; this difference defines the asymmetric actor--critic. Only the actor deploys.

Fine-tuning freezes the BC encoder and normalization statistics, then updates the head with narrow exploration and no entropy bonus. RL from scratch trains the full architecture with broader exploration and an entropy bonus. The different recipes suit each strategy, so the comparison between fine-tuning and RL from scratch does not isolate the effect of BC initialization.

Both use 40 parallel environments on one 48~GB-class cloud GPU per run (NVIDIA A40 or comparable), generalized advantage estimation with trace parameter $\lambda=0.95$, four minibatches per update, and gradient-norm clipping at 1.0. Fine-tuning uses policy-ratio clip $\epsilon=0.1$, a fixed learning rate of $5\times10^{-5}$, four epochs, eight rollout cycles per environment, entropy coefficient $\beta=0$, and initial depth and yaw scales $\sigma_\delta=\sigma_q=0.05$. RL from scratch uses $\epsilon=0.2$, an initial learning rate of $3\times10^{-4}$, five epochs, sixteen rollout cycles per environment, $\beta=0.01$, and $\sigma_\delta=\sigma_q=0.15$.

\section{EXPERIMENTAL SETUP}
\label{sec:setup}
\subsection{Physical system and observations}
\label{sec:execution}
A common finite state machine executes every target. \texttt{GAZE} holds a settled viewing pose; \texttt{HOVER} moves above the target; \texttt{ALIGN} sets yaw; \texttt{DESCEND} lowers the grapple; \texttt{CLOSE} closes the tongs; and \texttt{LIFT} raises the bundle. Inverse kinematics supplies joint setpoints to the programmable logic controller. The crane then transports at constant height and releases above the camera-measured fill height of the trailer compartment; these engineered motions deposit a stable, well-aligned bundle reliably. A cycle succeeds only after depositing at least one log. Simulation assesses and removes held logs after the lift.

\label{sec:observation}
The action box covers the rack's $2.0\times7.0$~m footprint and 1.4~m height; commands are clamped at the rack floor. A mast-mounted Stereolabs ZED~X stereo camera\footnote{\url{https://www.stereolabs.com/products/zed-x}} supplies depth, and simulation matches its resolution and projection. Each stream becomes a crane-frame cloud cropped to the action box plus a 0.5~m horizontal margin and sampled to 2048 points. Clouds with fewer points are zero-padded and masked; the margin exposes rails, poles, and ground.

\begin{figure}[t]
\centering
{\setlength{\fboxsep}{0pt}\setlength{\fboxrule}{0.4pt}%
\fbox{\includegraphics[width=\dimexpr0.49\columnwidth-0.8pt\relax]{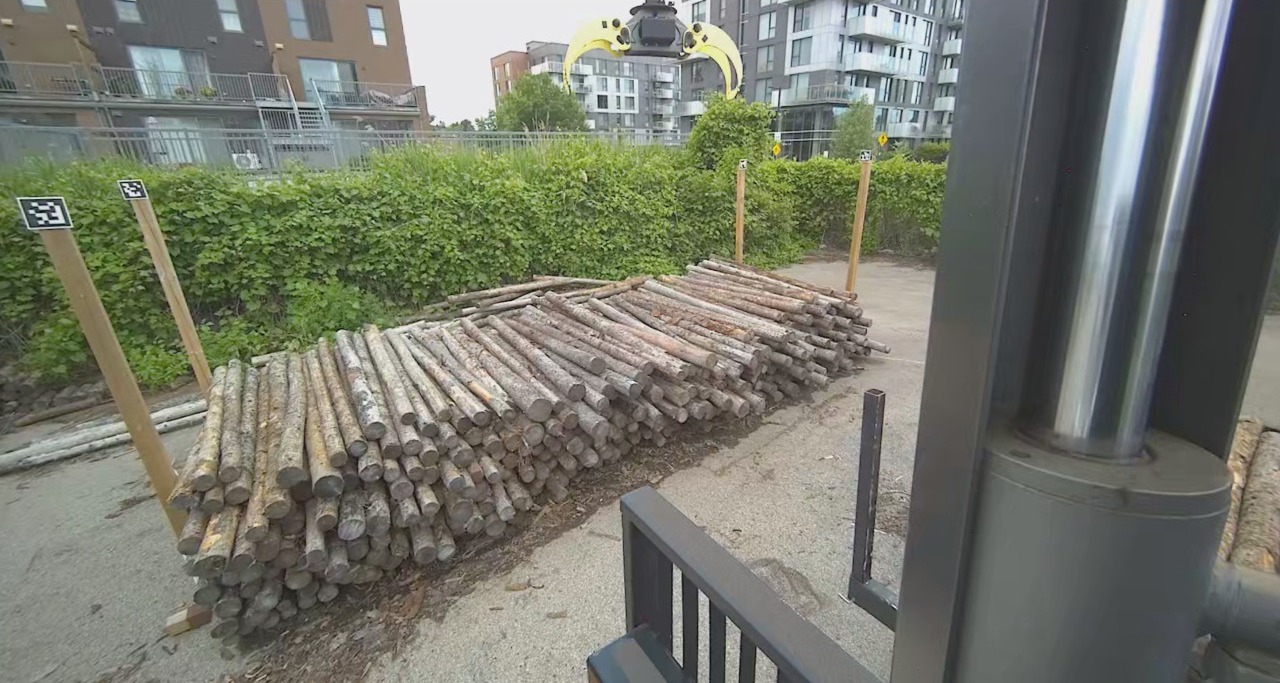}}\hfill
\fbox{\includegraphics[width=\dimexpr0.49\columnwidth-0.8pt\relax]{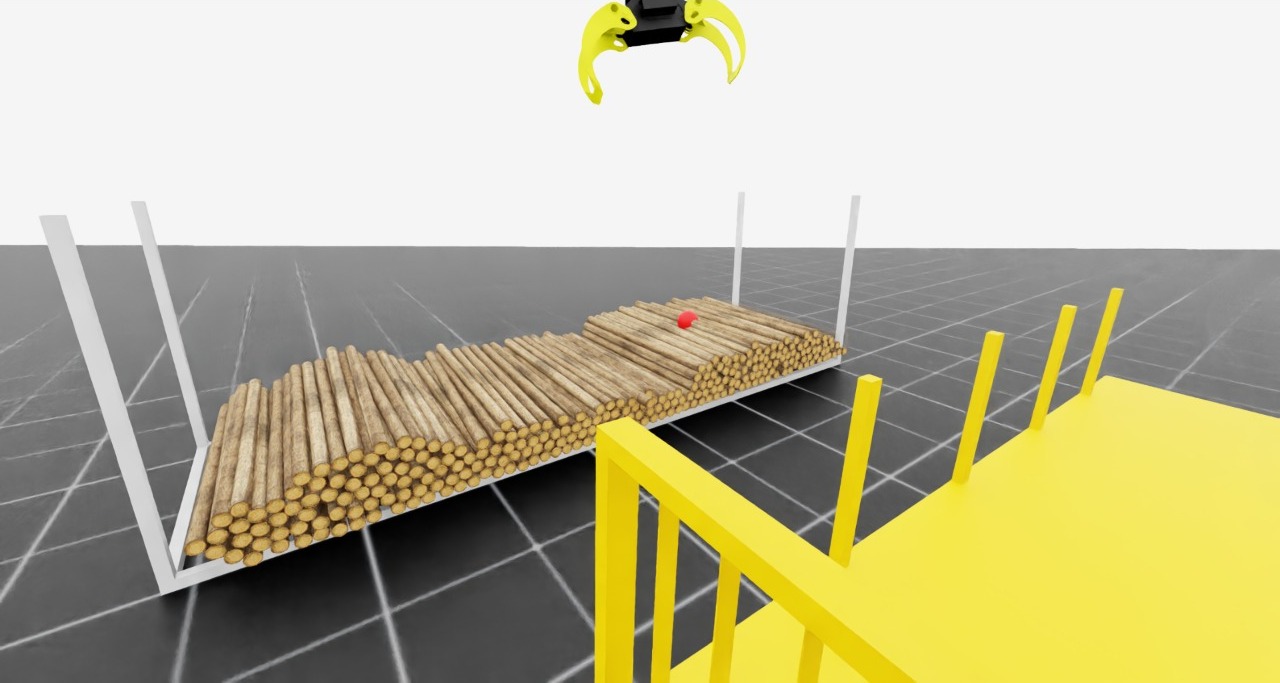}}}
\caption{Mast-camera views on the crane (left) and in simulation (right), using the calibrated camera and rack placement.}
\label{fig:gaze}
\end{figure}

The camera mount is calibrated against the crane's own grapple: recorded joint angles place its CAD model by forward kinematics, and the mount transform is fitted to the observed grapple by iterative closest point~\cite{besl1992method} with a floor-levelness term; fiducial markers~\cite{garridojurado2014automatic} on the rack posts, refined on the post columns of the cloud, locate the rack. Simulation uses this viewing geometry (Fig.~\ref{fig:gaze}).

\subsection{Simulation and grasp outcomes}
\label{sec:outcomes}
Training uses Isaac Lab~\cite{mittal2025isaaclab}, built on Isaac Sim, chosen for GPU-parallel PhysX rigid-body dynamics of hundreds of contacting logs across many environments and for on-GPU depth rendering that matches the deployed camera. The crane model includes actuated arm joints, grapple yaw and tongs, and passive suspension. Log geometry follows ten physical measurements: mean diameter 0.113~m, length 2.45~m, and taper approximately 1~cm of diameter per meter. Mass follows mesh volume at a density of 850~kg/m$^3$ (green softwood)~\cite{fpl2021wood}. Six size variants populate each 200-log episode, settled under gravity into randomized flat, mound, double-mound, ramp, or jagged profiles. Log contacts use published wood-on-wood friction coefficients~\cite{avallone2006marks}. Contact and suspension parameters are fixed.

\begin{figure}[t]
\centering
\includegraphics[width=\columnwidth]{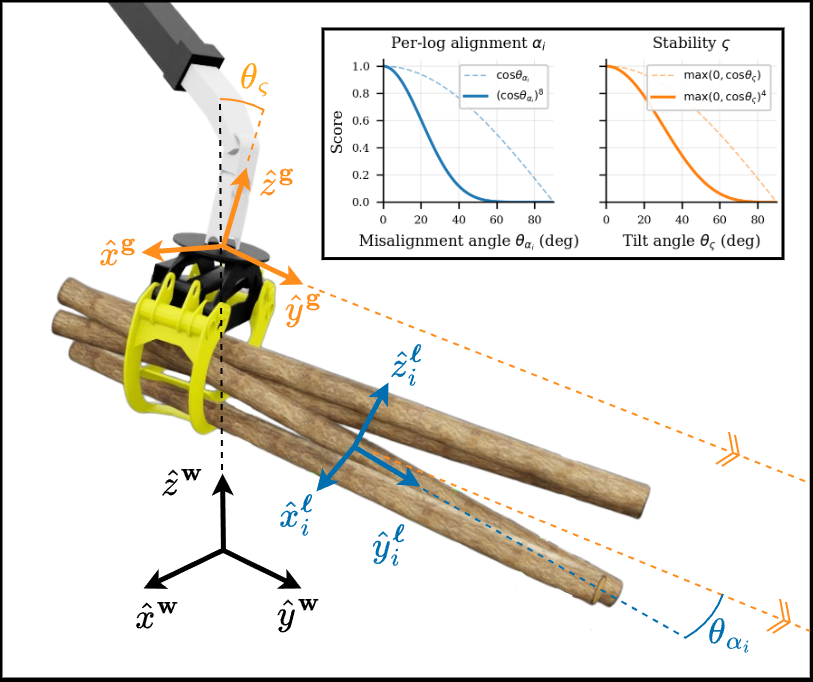}
\caption{Alignment and stability. Superscripts $w$, $g$, and $\ell$ denote world, grapple, and log frames; insets show the shaped scores in~\eqref{eq:outcomes}.}
\label{fig:outcomes}
\end{figure}

A simulated log is counted as held when its center is within 1.5~m of the grapple after lifting. For $n$ held logs, alignment $\alpha$ uses the misalignment angle $\theta_{\alpha_i}$ between the unit tong axis $\hat y^{\mathbf g}$, the grapple axis along which a held log lies when the tongs close squarely on it, and each unit log axis $\hat y^{\boldsymbol\ell}_i$; stability $\varsigma$ uses the grapple tilt angle $\theta_\varsigma$ (Fig.~\ref{fig:outcomes}):
\begin{equation}
 \alpha=\frac1n\sum_{i=1}^n\cos^8\theta_{\alpha_i},
 \qquad \varsigma=\max(0,\cos\theta_\varsigma)^4.
 \label{eq:outcomes}
\end{equation}
Here $\theta_\varsigma$ is the mean tilt over a one-second post-lift window, taken after settling or at the hold timeout. Both scores lie in $[0,1]$; the exponents sharpen them so that only nearly aligned logs and a nearly level load score close to one (Fig.~\ref{fig:outcomes}, insets).

\subsection{Compared policies and training}
\label{sec:baselines}
BC (\emph{scoring BC}), BC$\to$RL, and RL from scratch share the scoring head. The geometric heuristic is the rule-based baseline deployed on the crane. It applies the expert's top-of-pile strategy to the cloud: within a tight 1024-point crop with hand-placed pole exclusions, it takes the highest point in the middle half of the rack width that has three neighbors within 0.15~m, fits a local log axis for yaw or falls back to the rack direction, and grasps below that point at the expert's 0.25~m in simulation and a field-tuned 0.30~m on the crane.

Two alternative heads isolate the representation (Sec.~\ref{sec:heads}): under BC, a \emph{regression head} (\emph{regression BC}) maps the global feature $g$ to one box-normalized position and yaw by squared error, with the demonstrations, preprocessing, backbone, and optimizer of the scoring head unchanged; under RL from scratch, a \emph{Gaussian head} over grasp coordinates replaces categorical point selection, sharing environment and reward.

Fine-tuning and RL from scratch use three training seeds and 50k cycles per run. Checkpoints are screened over 40 deterministic episodes, and the one emptying most racks receives the 100-episode evaluation and deployment.

\subsection{Evaluation protocols}
\label{sec:sim_protocol}
In simulation, six policies run over 100 episodes on shared pile seeds from the training distribution. Episodes end at an empty rack or horizon $H=30$ cycles. Deterministic evaluation selects the highest-scoring point and mean depth and yaw. Full counts emptied racks; success requires lifting at least one log. Removed inventory excludes logs knocked out of bounds. The metric $c_{95}$ counts cycles until 95\% of the initial inventory has been removed. Its mean includes only episodes reaching that threshold: 99 for the expert, 46 for regression BC, and 100 for every other policy. Cycles to termination count grasp decisions until the episode ends.

\begin{figure*}[t]
\centering
{\setlength{\fboxsep}{0pt}\setlength{\fboxrule}{0.4pt}%
\includegraphics[width=0.322\textwidth]{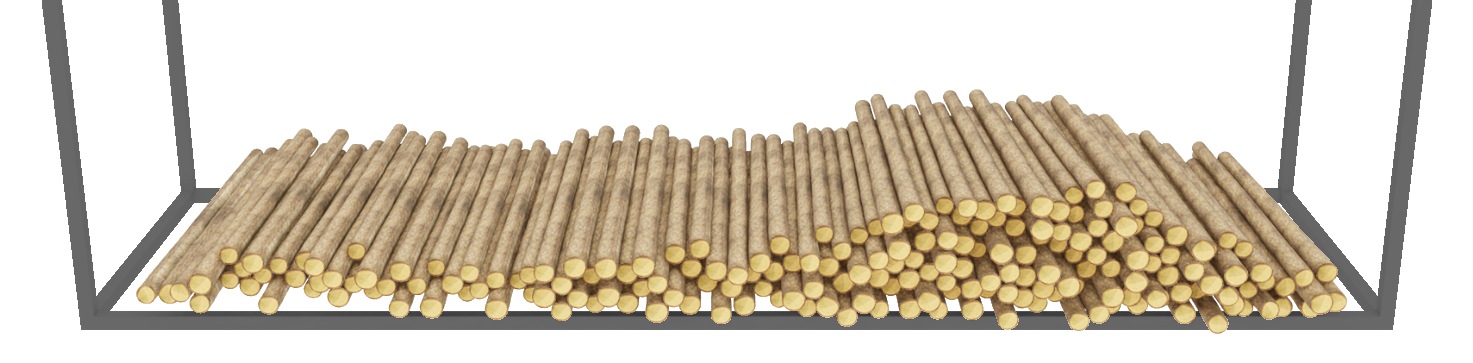}\hfill
\includegraphics[width=0.322\textwidth]{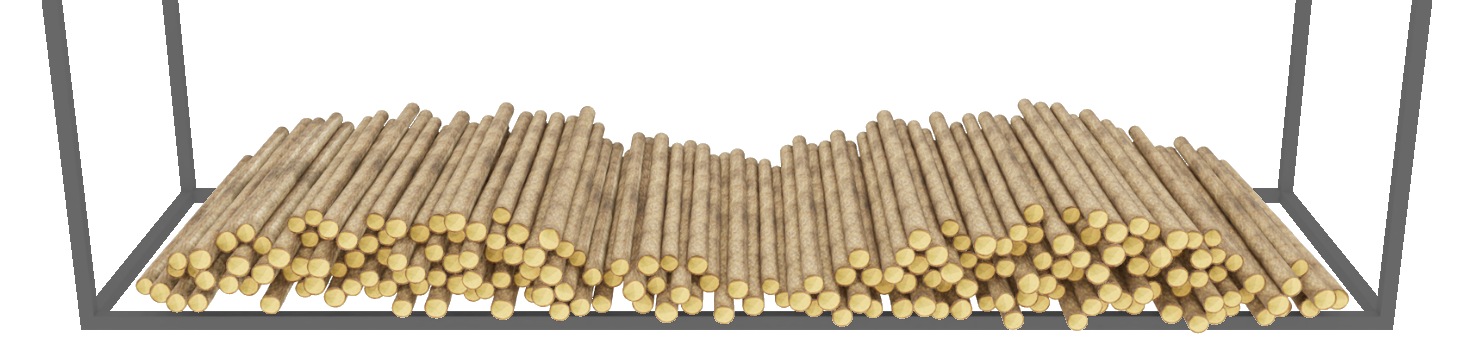}\hfill
\includegraphics[width=0.322\textwidth]{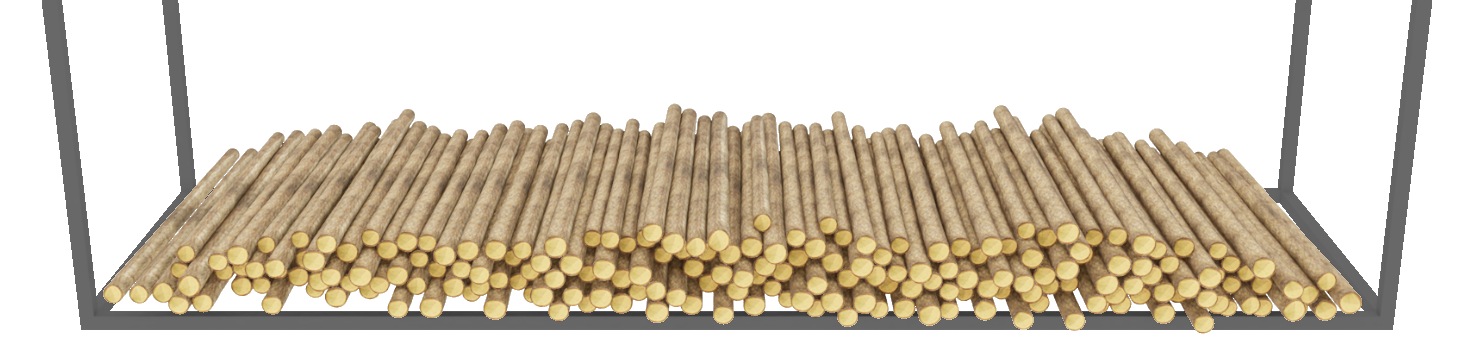}\\[1mm]
\fbox{\includegraphics[width=\dimexpr0.322\textwidth-0.8pt\relax]{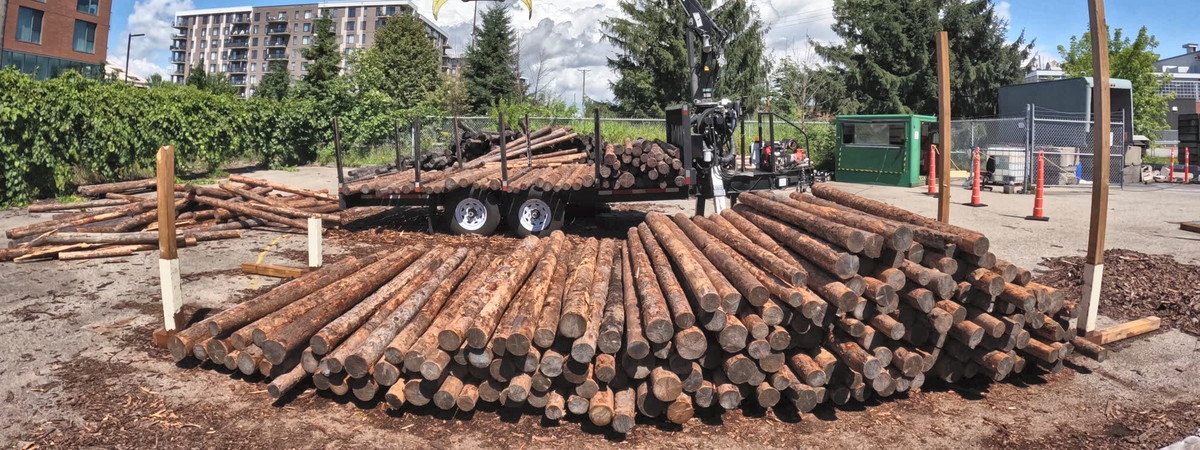}}\hfill
\fbox{\includegraphics[width=\dimexpr0.322\textwidth-0.8pt\relax]{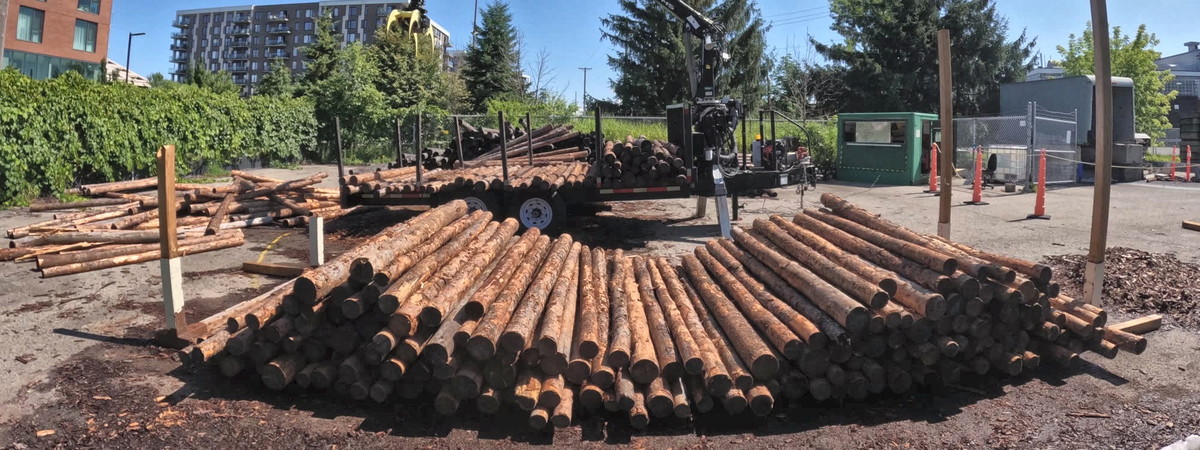}}\hfill
\fbox{\includegraphics[width=\dimexpr0.322\textwidth-0.8pt\relax]{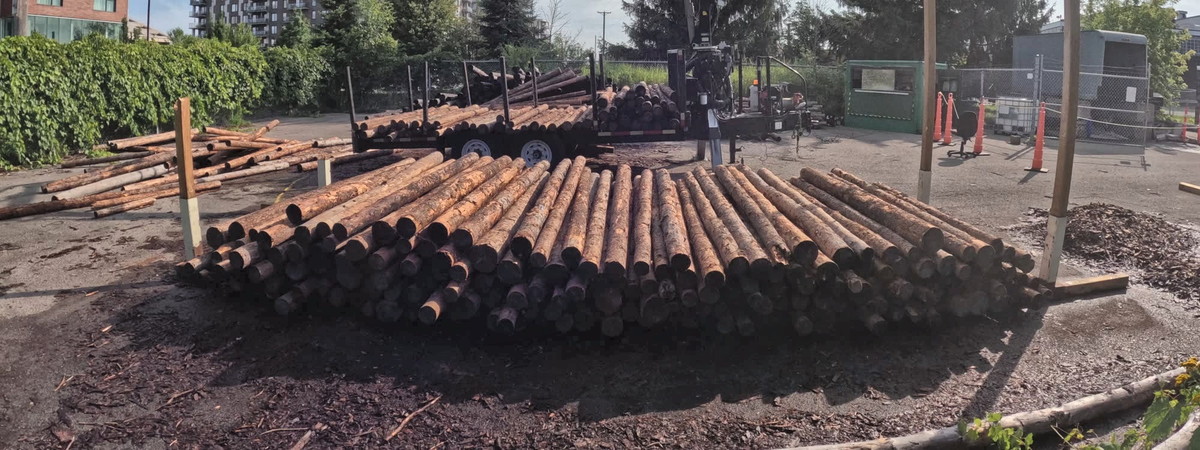}}}
\caption{Single mound, double mound, and flat rack in simulation (top) and before crane trials (bottom). Categories match, but each physical pile is rebuilt. The source rack is in front of the destination trailer.}
\label{fig:pile_shapes}
\end{figure*}

\label{sec:field_protocol}
On the physical crane testbed, the geometric heuristic, BC, BC$\to$RL, and RL from scratch each run once on a single mound, double mound, and flat rack, giving twelve trials. Figure~\ref{fig:pile_shapes} shows these shapes in simulation and on the testbed. Rebuilt piles share shape categories but are not identical paired states. Reconstructed initial inventories range from 199 to 219 logs. Analysis ends at the first empty rack, three consecutive empty cycles, or 30 cycles. Results for RL from scratch include manual depth offsets (Sec.~\ref{sec:depth}).

Clear is the share of initial inventory deposited in the trailer; knocked-off logs are not cleared. An empty rack can therefore have Clear below 100\%. Here $c_{95}$ counts cycles until 95\% is deposited, and $n$, logs per grasp, is the mean over a trial's successful cycles. Because a closed grapple hides individual poses, recordings score alignment from drooping or falling logs (1), poor (2), acceptable (3), good (4), to a compact, well-aligned bundle (5). Pooled rows treat the three trials as one campaign: Succ.\ and Clear are ratios of the summed counts, Cycles is the total, and $n$, alignment, and $\varsigma$ average over all successful cycles; $c_{95}$ is per trial only.

\section{RESULTS}
\label{sec:results}
\subsection{Imitation, exploration, and simulated clearing}
\label{sec:simulation}
\label{sec:tradeoffs}
\begin{figure}[t]
\centering
\includegraphics[width=\columnwidth]{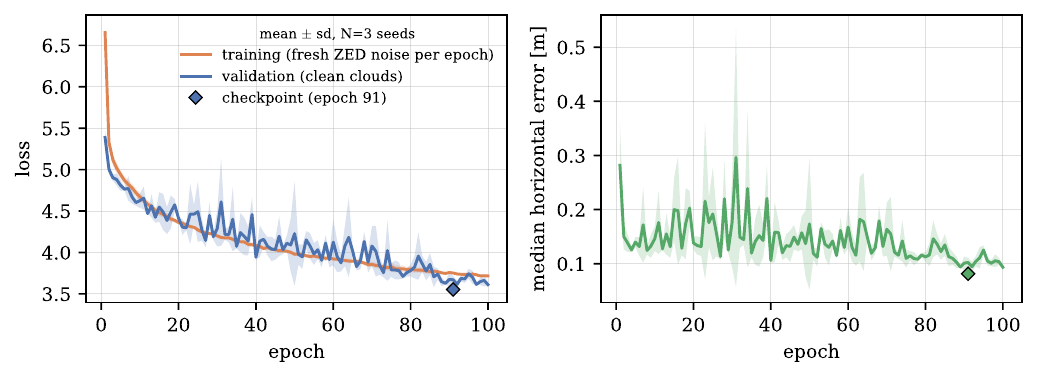}
\caption{BC loss (left) and validation grasp error (right), the median horizontal distance from the decoded grasp position to the expert target, mean and standard deviation over three seeds. Training uses fresh camera noise; validation is clean. Diamonds mark the deployed checkpoint.}
\label{fig:bccurves}
\end{figure}

\begin{figure}[t]
\centering
\includegraphics[width=\columnwidth]{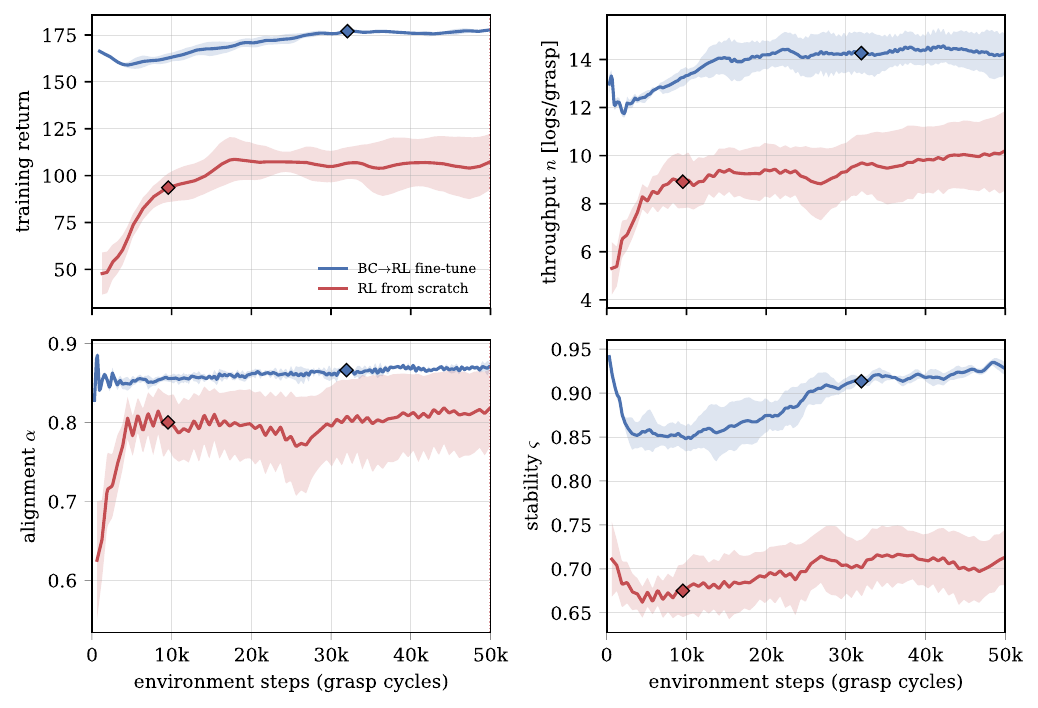}
\caption{RL return and handling, mean and standard deviation over three seeds. Training samples actions; evaluation uses the highest score and predicted depth and yaw. Diamonds mark checkpoints.}
\label{fig:rlcurves}
\end{figure}

BC and its fine-tuned policy achieve the highest full-clearing rates on the shared simulated piles (Table~\ref{tab:sim}). BC training reduces loss and median horizontal grasp error (Fig.~\ref{fig:bccurves}). In training, fine-tuning reaches higher mean return and handling than learning from scratch (Fig.~\ref{fig:rlcurves}).

\begin{table}[!t]
\centering
\caption{Simulation on 100 shared piles (mean $\pm$ standard deviation). $n$: logs per successful grasp; $\alpha$: alignment; $\varsigma$: stability; Succ.: successful-cycle rate; Full: emptied-rack rate; $c_{95}$: cycles to 95\% removal in episodes reaching it.}
\label{tab:sim}
\scriptsize
\renewcommand{\arraystretch}{0.9}
\setlength{\tabcolsep}{2.5pt}
\resizebox{\columnwidth}{!}{%
\begin{tabular}{lccccccc}
\toprule
Method & $n$ & $\alpha$ & $\varsigma$ & Succ.\,(\%) & Full\,(\%) & $c_{95}$ & Cycles \\
\midrule
Privileged expert & \statcell{11.51}{1.18} & \statcell{0.977}{0.010} & \statcell{0.948}{0.016} & \statcell{76.3}{16.3} & 63 & \statcell{16.1}{1.7} & \statcell{23.9}{5.2} \\
Geometric heuristic & \statcell{11.23}{1.05} & \statcell{0.974}{0.010} & \statcell{0.931}{0.017} & \statcell{79.8}{9.9} & 89 & \statcell{16.9}{1.6} & \statcell{22.9}{3.7} \\
\midrule
RL from scratch & \statcell{11.01}{1.15} & \statcell{0.950}{0.016} & \statcell{0.862}{0.020} & \statcell{81.3}{9.6} & 87 & \statcell{18.3}{2.1} & \statcell{23.0}{4.1} \\
BC, regression head & \statcell{10.65}{1.92} & \statcell{0.979}{0.011} & \statcell{0.951}{0.015} & \statcell{60.8}{19.8} & 33 & \statcell{20.4}{4.6} & \statcell{27.5}{4.1} \\
BC, scoring head & \statcell{11.77}{1.00} & \statcell{0.980}{0.008} & \statcell{0.920}{0.024} & \statcell{91.1}{6.2} & 98 & \statcell{15.2}{1.4} & \statcell{18.9}{2.5} \\
BC$\to$RL & \statcell{11.89}{0.81} & \statcell{0.984}{0.006} & \statcell{0.955}{0.017} & \statcell{88.3}{5.6} & 99 & \statcell{16.2}{1.3} & \statcell{19.2}{1.9} \\
\bottomrule
\end{tabular}}
\end{table}

Scoring BC finishes more racks than its privileged expert. Unfinished expert episodes retain a median of two logs, typically isolated against the floor. The common closing sweep can miss the same log repeatedly, leaving the height rule to select it again. BC learns only from successful expert cycles, which may explain its stronger completion.

Fine-tuning changes handling more clearly than completion. It clears two paired piles that BC misses and misses one that BC clears. Stability rises by $0.035\pm0.003$ (paired mean $\pm$ standard error), knocked-off logs approximately halve, and cycle success falls $2.8\pm0.8$ percentage points. The reward permits this trade-off between grasping more often and carrying a more stable load. The $c_{95}$ and termination columns separate bulk removal from clearing the last remaining logs.

\begin{figure}[t]
\centering
\includegraphics[width=\columnwidth]{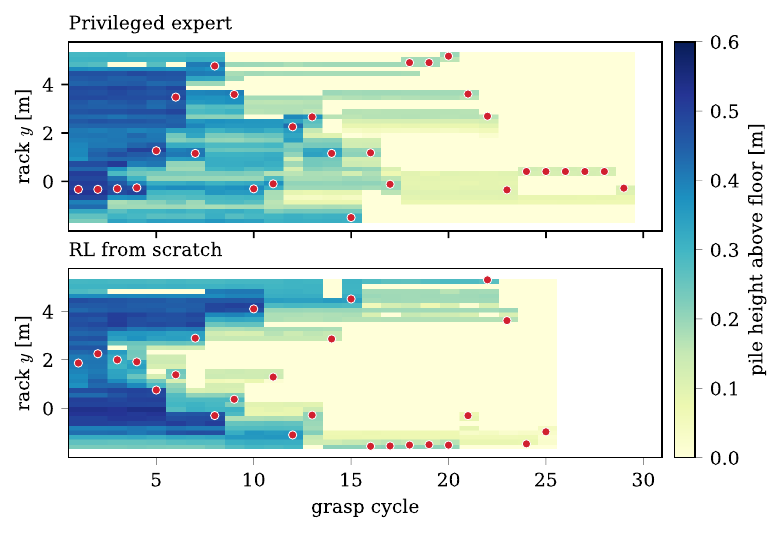}
\caption{Targets from the expert and RL from scratch on one shared pile. Red points mark targets by cycle; color gives remaining height. Reported spacing medians use all episodes.}
\label{fig:strategy}
\end{figure}

Figure~\ref{fig:strategy} illustrates the regional strategy discovered by RL from scratch, which works one area down before moving on. Consecutive targets in the first five cycles are a median 0.4~m apart, against 1.5~m for the expert and fine-tuned policy, and its targets remain farther from the highest point over complete episodes. Exploration departs from the demonstrated strategy, but under the tested reward and budget it does not improve completion over BC and has the weakest handling.

\subsection{Representation and design ablations}
\label{sec:heads}
The controlled BC comparison tests whether learning a score at each observed point improves on directly predicting grasp coordinates. Across three training seeds, the scoring head empties 98, 98, and 99 of 100 racks; the regression head empties 33, 0, and 0. For the regression run in Table~\ref{tab:sim}, unfinished episodes retain a median 45 logs, so regression fails while much of the pile remains, well before the difficult final remainder.

Squared-error regression can average alternative expert targets. Between two similar mounds, this produces a target with no material beneath it. Across 1784 recorded decisions from three regression runs, 14.6--24.7\% have no observed point within 0.5~m horizontally; 5.1--8.6\% reissue such a target within 0.15~m of the previous one. Scoring BC has neither diagnostic failure over 2140 decisions.

\begin{figure}[t]
\centering
\includegraphics[width=\columnwidth]{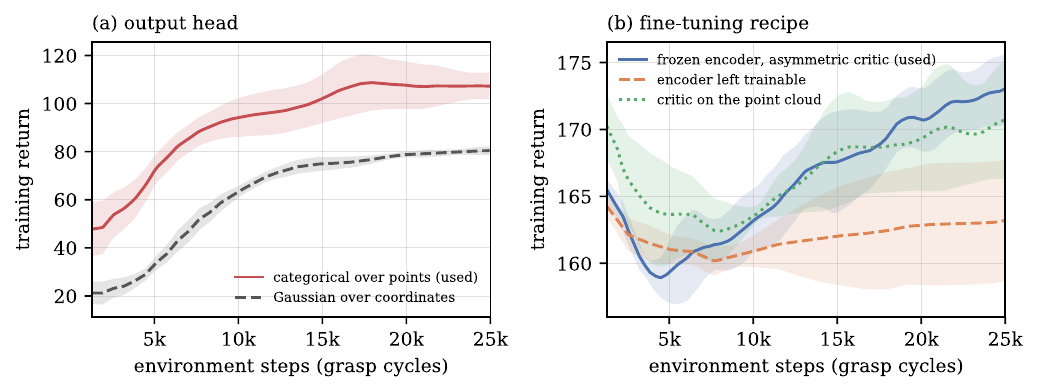}
\caption{Training return over three seeds: (a) categorical and Gaussian coordinate exploration; (b) encoder and critic variants during fine-tuning. Curves share reward within each panel.}
\label{fig:abl}
\end{figure}

A separate three-seed, 25k-cycle comparison of RL from scratch holds the environment and reward fixed while changing categorical point selection to Gaussian coordinate exploration. Table~\ref{tab:abl} separates sampled-action training performance from deterministic completion at each seed's best screened checkpoint; Fig.~\ref{fig:abl} shows return. The categorical head's screened completion rates are 97.5\%, 57.5\%, and 0\%; the Gaussian head empties no rack on any seed. This supports exploring only observed points, although learning from scratch varies widely by seed.

\begin{table}[t]
\centering
\caption{Three-seed ablations (mean $\pm$ standard deviation). Clear and handling use sampled cycles 12.5k--25k; Argmax full uses the best 40-episode deterministic screen per seed.}
\label{tab:abl}
\scriptsize
\renewcommand{\arraystretch}{0.9}
\setlength{\tabcolsep}{3pt}
\resizebox{\columnwidth}{!}{%
\input{tables/ablations}}
\end{table}

\begin{figure}[t]
\centering
\includegraphics[width=\columnwidth]{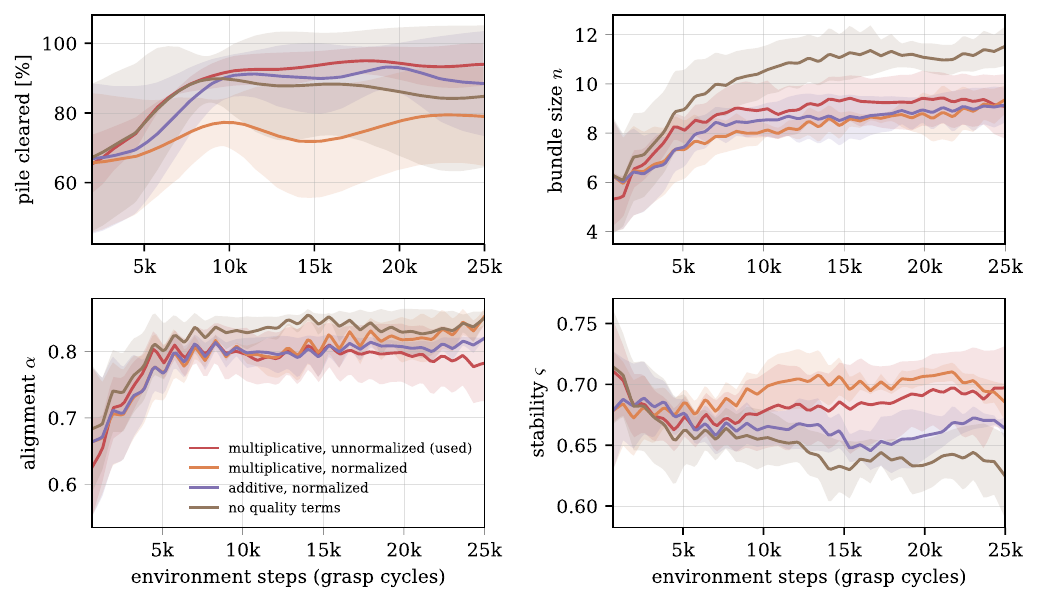}
\caption{Reward ablation, RL from scratch: clearing, bundle size, alignment, and stability, mean $\pm$ standard deviation over three seeds.}
\label{fig:reward_abl}
\end{figure}

The reward ablation compares~\eqref{eq:reward} with three alternatives for successful cycles: normalized multiplication $10f\alpha\varsigma$ and normalized addition $f+\alpha+\varsigma$, which reward taking what the target offers instead of seeking the tallest stack, and log count $n$, which drops the quality terms. Here $f=\min(n/\tilde n,1)$ and $\tilde n$ is the number of logs stacked at the target, capped at 15. Every variant keeps the empty-cycle penalty. The deployed reward gives the highest mean training clearing; removing quality terms increases bundle size and lowers stability (Table~\ref{tab:abl}, Fig.~\ref{fig:reward_abl}).

The fine-tuning ablations change either encoder freezing or the critic's observation. Leaving the encoder trainable lowers training return and widens the observed spread in screened completion, with rates from 72.5\% to 100\%. With the encoder frozen, the critic given privileged log poses finishes with higher mean training return and a narrower late seed band than the critic on the actor's point cloud (Fig.~\ref{fig:abl}).

\begin{figure}[t]
\centering
\includegraphics[width=0.49\columnwidth]{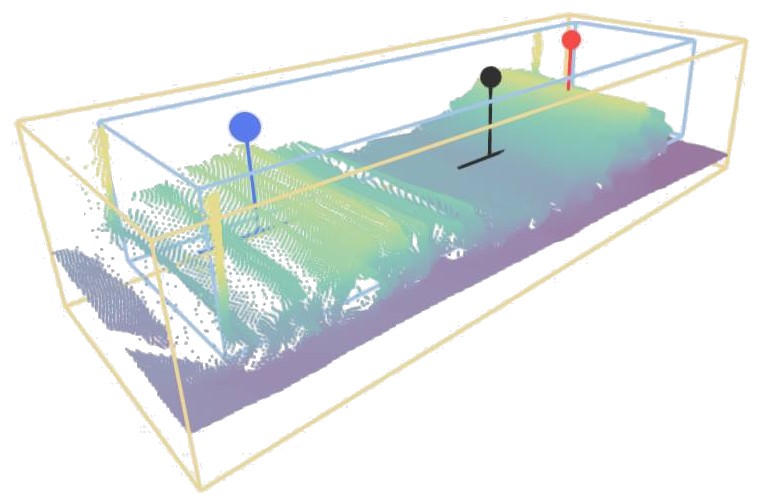}\hfill
\includegraphics[width=0.49\columnwidth]{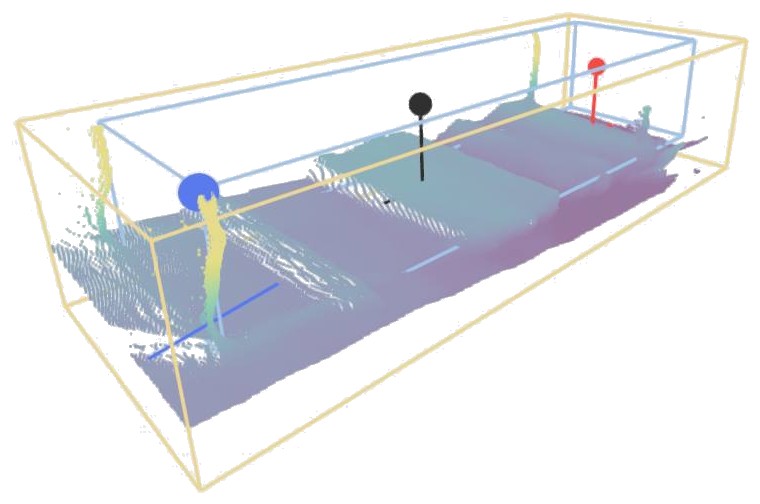}
\caption{Offline targets on full (left) and depleted (right) clouds, colored by height: regression (black), margin-trained scoring (red), tight-crop scoring (blue). Inner box: tight rack crop; outer box: 0.5~m margin crop.}
\label{fig:replay}
\end{figure}

Offline replay compares regression BC with two scoring variants on recorded crane clouds (Fig.~\ref{fig:replay}). One variant was trained on tight rack crops, the other on the 0.5~m margin crop that exposes surrounding structure. Regression again predicts intermediate targets between mounds, while margin-trained scoring chooses a mound point. Tight-crop scoring repeatedly selects rails and poles it never saw in training. This motivated deploying the margin-trained policy.

\begin{figure}[!b]
\centering
\includegraphics[width=\columnwidth]{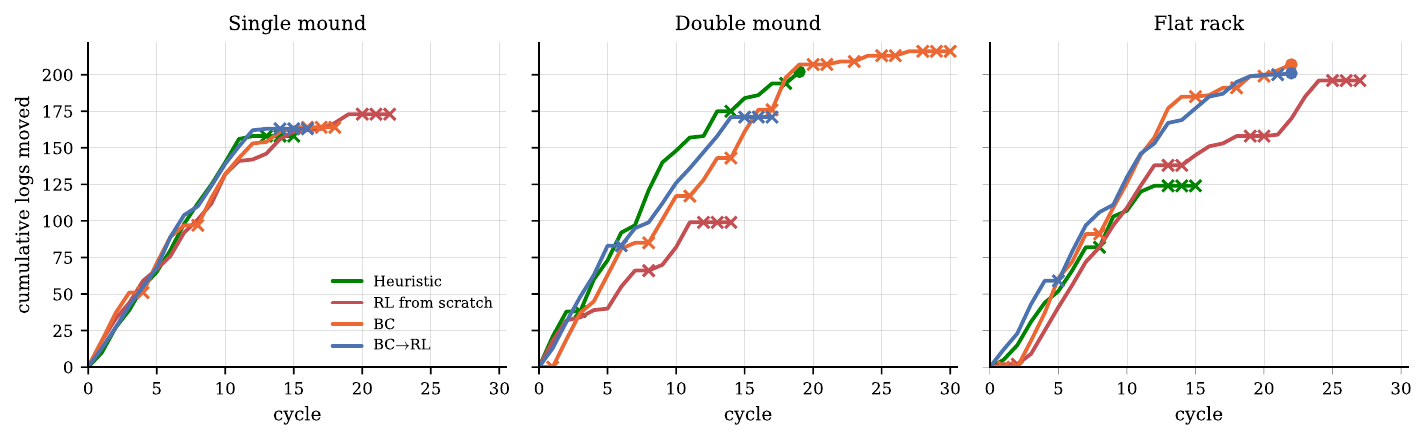}
\caption{Cumulative logs moved in each field trial. Crosses mark empty cycles, dots a cleared rack; RL from scratch includes depth offsets.}
\label{fig:shape_curves}
\end{figure}

\subsection{Clearing and handling on the crane testbed}
\label{sec:field}
\label{sec:field_results}
\begin{figure}[t]
\centering
\includegraphics[width=\columnwidth]{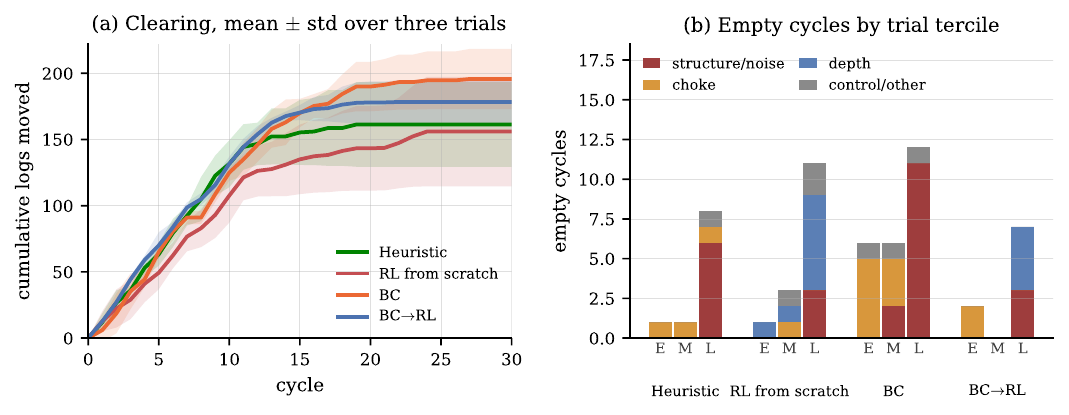}
\caption{Field campaign: (a) cumulative logs moved, mean $\pm$ standard deviation over one trial per shape, ended trials holding their final value; (b) empty-cycle causes by early (E), middle (M), and late (L) thirds.}
\label{fig:curves}
\end{figure}

BC recovers the most pooled inventory; fine-tuning gives the highest cycle success (Table~\ref{tab:field}). Figure~\ref{fig:shape_curves} gives each trial; Fig.~\ref{fig:curves} the pooled progression and empty-cycle causes.

\begin{table}[!t]
\centering
\caption{Twelve crane trials. $n$: logs per depositing cycle; Align.: 1--5 bundle score; $\varsigma$: stability over valid successful cycles; these three are mean $\pm$ standard deviation over the trial's successful cycles; Succ.: depositing-cycle rate; Clear: deposited inventory. RL from scratch includes depth offsets.}
\label{tab:field}
\scriptsize
\renewcommand{\arraystretch}{0.88}
\setlength{\tabcolsep}{3pt}
\resizebox{\columnwidth}{!}{%
\begin{tabular}{lccccccc}
\toprule
Policy & $n$ & Align. & $\varsigma$ & Succ.\,(\%) & Clear\,(\%) & $c_{95}$ & Cycles \\
\midrule
\multicolumn{8}{l}{\emph{Single mound}} \\
Heuristic & $13.2 \pm 4.2$ & $4.42 \pm 0.95$ & $0.947 \pm 0.049$ & 80.0 & 79.0 & -- & 15 \\
RL from scratch & $9.1 \pm 5.6$ & $4.05 \pm 0.89$ & $0.781 \pm 0.116$ & 86.4 & 86.5 & -- & 22 \\
BC & $12.6 \pm 6.2$ & $4.23 \pm 0.97$ & $0.942 \pm 0.033$ & 72.2 & 82.4 & -- & 18 \\
BC$\to$RL & $12.5 \pm 4.6$ & $4.08 \pm 1.07$ & $0.918 \pm 0.059$ & 81.2 & 81.5 & -- & 16 \\
\midrule
\multicolumn{8}{l}{\emph{Double mound}} \\
Heuristic & $12.6 \pm 7.1$ & $4.44 \pm 1.06$ & $0.937 \pm 0.032$ & 84.2 & 100.0 & 17 & 19 \\
RL from scratch & $9.9 \pm 6.0$ & $2.90 \pm 1.45$ & $0.777 \pm 0.155$ & 71.4 & 49.5 & -- & 14 \\
BC & $12.7 \pm 6.3$ & $4.00 \pm 1.37$ & $0.933 \pm 0.055$ & 56.7 & 98.6 & 22 & 30 \\
BC$\to$RL & $13.2 \pm 3.9$ & $4.15 \pm 0.95$ & $0.898 \pm 0.039$ & 76.5 & 85.5 & -- & 17 \\
\midrule
\multicolumn{8}{l}{\emph{Flat rack}} \\
Heuristic & $11.3 \pm 5.3$ & $4.36 \pm 0.77$ & $0.969 \pm 0.018$ & 73.3 & 62.0 & -- & 15 \\
RL from scratch & $10.3 \pm 5.1$ & $3.56 \pm 1.17$ & $0.732 \pm 0.132$ & 70.4 & 98.0 & 24 & 27 \\
BC & $12.9 \pm 6.9$ & $4.19 \pm 0.88$ & $0.952 \pm 0.038$ & 72.7 & 99.5 & 19 & 22 \\
BC$\to$RL & $10.1 \pm 6.4$ & $4.10 \pm 1.26$ & $0.894 \pm 0.083$ & 90.9 & 99.5 & 18 & 22 \\
\midrule
\multicolumn{8}{l}{\emph{Pooled}} \\
Heuristic & $12.4 \pm 5.9$ & $4.41 \pm 0.95$ & $0.949 \pm 0.038$ & 79.6 & 80.4 & -- & 49 \\
RL from scratch & $9.8 \pm 5.5$ & $3.62 \pm 1.21$ & $0.763 \pm 0.133$ & 76.2 & 78.0 & -- & 63 \\
BC & $12.8 \pm 6.5$ & $4.13 \pm 1.11$ & $0.942 \pm 0.044$ & 65.7 & 93.8 & -- & 70 \\
BC$\to$RL & $11.6 \pm 5.5$ & $4.11 \pm 1.13$ & $0.901 \pm 0.067$ & 83.6 & 88.9 & -- & 55 \\
\bottomrule
\end{tabular}}
\end{table}

BC and BC$\to$RL deposit more inventory than the geometric heuristic on two shapes and over the pooled campaign. The flat rack makes the difference clearest: both learned policies empty it, while the heuristic stops at 62.0\% Clear. The double mound reverses it: the heuristic deposits everything.

RL from scratch falls behind early on the double mound. A matched first-fifteen-cycle comparison, reached by every heuristic, BC, and fine-tuned trial, gives success rates of 80\%, 78\%, and 89\%, respectively. The heuristic and fine-tuned rates are within two percentage points of their simulated rates. This early-window agreement does not establish that later clearing or handling transfers equally well.

The fine-tuned policy's simulated stability advantage does not transfer: its stability is below BC and the heuristic on every shape, by 0.02 to 0.06, about the size of its simulated gain. RL from scratch has the lowest stability on all three shapes, with the lowest pooled alignment and logs per successful cycle. Observed handling failures include grasps near log ends and loss of part of the bundle during lifting.

\subsection{Where physical clearing fails}
\label{sec:failures}
Empty-cycle causes distinguish structure or noise targeting, closures blocked by too much material in the tongs (\emph{chokes}), commands that close above material (\emph{depth faults}), and control faults. Figure~\ref{fig:curves}(b) groups these causes by trial third. All six heuristic structure/noise cycles occur in the final third. Pole returns survive the heuristic's hand-placed, axis-aligned exclusion boxes because the poles are thick and the boxes sit on a calibration that drifts, and the cloud also carries stray noise returns. Once such points stand above the remaining logs, the height rule repeatedly chooses them; an empty grasp leaves little change to prompt a different target.

Learning does not stop structure targeting, but it ends the trap. BC has 13 structure/noise cycles in 70, against six in 49 for the heuristic. On the flat rack, however, BC returns to material after each of its three structure cycles and finishes. BC$\to$RL has three structure cycles in 55, all on the base rails, the horizontal members at the rack ends, and never a pole. These match logs in diameter and height; the rack-pole penalty does not cover them. The real rack exposes ground between its base rails; the simulated rack has a solid floor at rail height.

Failures remain even when the target lands on logs. In its 30-cycle double-mound trial, BC has no gap-targeting cycle, yet produces four chokes on the steep mound, one control fault, and eight late structure cycles. The choked commands lie within the successful depth band of the campaign. On a steep flank, that depth can enclose more material than on a level pile, blocking closure; simulation shows no chokes.

\label{sec:depth}
BC$\to$RL completes thirteen successful single-mound cycles before closing above remaining logs on one side of the rack. The trial ends in that empty-cycle run. Residual camera and kinematic errors are plausible contributors, but this trial does not separate them from the policy's depth prediction.

RL from scratch shows a systematic depth bias: with applied offsets removed, its mound targets lie a median 0.17~m above the observed surface. BC and BC$\to$RL learn depth from demonstrations relabeled using crane commissioning; RL from scratch learns it only in simulation. The mound trials use a 0.15~m downward offset and still grasp near the surface. On the flat rack, the offset follows two above-surface closes and is increased later, giving a median depth of 0.21~m below the surface, within the other trials' 0.18--0.33~m band. Its 98.0\% Clear includes that adaptation, and it also deposits the most inventory on the single mound.

\section{DISCUSSION}
\label{sec:discussion}
BC and BC$\to$RL recover more inventory than the geometric heuristic on two pile shapes and across the pooled campaign while using a wider cloud without a geometric support filter or separate segmentation network. Structure selection still causes failures, particularly for BC, but on the flat rack BC returns to material after each structure-targeting cycle, while the heuristic's repeated selection of residual structure ends its trial. Both were trained entirely in simulation and run unchanged on the crane, and still cleared more than the heuristic.

Neither RL recipe yields one best physical policy: BC recovers the most pooled inventory, and BC$\to$RL has the highest pooled cycle success but loses its simulated stability advantage on the crane testbed.

Beyond log handling, the results motivate evaluating task completion and handling separately over the full depletion of a pile: better simulated handling need not yield better physical handling.

The evidence has limits: one crane and rack, one rebuilt pile per policy and shape. Three trials per policy cannot establish cross-platform performance or identify the cause of the stability result, although the ordering held on every shape.

Simulation misses physical chokes and the field stability ordering. Possible sources include the crane's hydraulic and suspension dynamics, unmeasured friction, and the bark and knots of real logs, which the smooth rigid assets lack. Future work should measure how transfer depends on these contact parameters, correct the remaining depth faults with small amounts of real-trial data, extend the rack-pole penalty to the base rails, and randomize contact properties during training.

\section{CONCLUSION}
One point-selection, depth, and yaw network serves BC, RL, and deployment: it learns in simulation and executes complete crane cycles with unchanged weights. BC and BC$\to$RL recover more inventory than the geometric heuristic overall with less hand-tuned filtering. Fine-tuning's stability gain did not carry over to the crane testbed: handling is where the sim-to-real gap remains, and closing it, with real-trial data or more accurate crane dynamics modeling, is the next step.

\FloatBarrier
\bibliographystyle{IEEEtran}
\bibliography{references}
\end{document}

%% file: tables/ablations.tex
\begin{tabular}{lcccc|c}
\toprule
Variant & Clear\,(\%) & $n$ & $\alpha$ & $\varsigma$ & Argmax full\,(\%) \\
\midrule
\multicolumn{6}{l}{\emph{Output head}} \\
Categorical over points (used) & \statcell{94.0}{4.8} & \statcell{9.3}{1.0} & \statcell{0.79}{0.04} & \statcell{0.69}{0.03} & \statcell{51.7}{40.0} \\
Gaussian over coordinates      & \statcell{90.6}{1.2} & \statcell{9.9}{0.1} & \statcell{0.84}{0.01} & \statcell{0.65}{0.01} & \statcell{0.0}{0.0} \\
\midrule
\multicolumn{6}{l}{\emph{Reward}} \\
Multiplicative, unnormalized (used) & \statcell{94.0}{4.8} & \statcell{9.3}{1.0} & \statcell{0.79}{0.04} & \statcell{0.69}{0.03} & \statcell{51.7}{40.0} \\
Multiplicative, normalized          & \statcell{76.4}{14.5} & \statcell{8.8}{0.6} & \statcell{0.83}{0.02} & \statcell{0.70}{0.00} & \statcell{5.8}{5.1} \\
No quality terms                    & \statcell{86.2}{18.0} & \statcell{11.2}{0.7} & \statcell{0.84}{0.02} & \statcell{0.63}{0.02} & \statcell{30.8}{38.3} \\
Additive, normalized                & \statcell{90.4}{10.6} & \statcell{8.9}{0.7} & \statcell{0.81}{0.01} & \statcell{0.66}{0.01} & \statcell{13.3}{12.5} \\
\midrule
\multicolumn{6}{l}{\emph{Fine-tuning recipe}} \\
Frozen encoder, asymmetric critic (used) & \statcell{100.0}{0.0} & \statcell{14.1}{0.7} & \statcell{0.86}{0.00} & \statcell{0.87}{0.02} & \statcell{96.7}{2.4} \\
Encoder left trainable                   & \statcell{99.9}{0.0}  & \statcell{13.9}{0.8} & \statcell{0.87}{0.00} & \statcell{0.86}{0.03} & \statcell{90.0}{12.4} \\
Critic on the point cloud                & \statcell{100.0}{0.0} & \statcell{13.7}{0.6} & \statcell{0.86}{0.01} & \statcell{0.88}{0.03} & \statcell{95.0}{7.1} \\
\bottomrule
\end{tabular}